\documentclass[runningheads, a4paper]{llncs}
\usepackage[T1]{fontenc}
\usepackage{graphicx}
\usepackage{siunitx}
\usepackage{color}
\usepackage{hyperref}

\providecommand{\IfDocumentMetadataTF}[2]{#2}
\IfDocumentMetadataTF{}{\usepackage{accsupp}}
\makeatletter
\IfDocumentMetadataTF{}{%
  \newcommand*{\PublicationEndAltText}{\EndAccSupp{}}%
  \define@key{Gin}{alt}{%
    \BeginAccSupp{method=pdfstringdef,unicode,Alt={#1}}%
    \aftergroup\PublicationEndAltText
  }%
}
\makeatother

\providecommand{\xmpquote}[1]{#1}
\makeatletter
\@ifundefined{xmpquote}{\newcommand{\xmpquote}[1]{#1}}{}
\makeatother
\hypersetup{
  pdftitle={Teaching a Minimalist Machine to Discover Recursive Programs for Arithmetic},
  pdfauthor={Dominik Magiera; Christiane Wiebel-Herboth; Frank Jäkel},
  pdflang={en-US},
  pdfkeywords={\xmpquote{Inductive Program Synthesis}, \xmpquote{Curriculum Learning}, \xmpquote{Recursive Programs}},
  hidelinks,
  pdfcreator={LaTeX via pandoc}}

\begin{document}
\title{Teaching a Minimalist Machine to Discover Recursive Programs for Arithmetic}
\titlerunning{Teaching a Minimalist Machine}
\author{%
Dominik Magiera\inst{1}\orcidID{0009-0002-0643-9973}%
 \and
Christiane Wiebel-Herboth\inst{2}\orcidID{0000-0002-6574-3820}%
 \and
Frank Jäkel\inst{1}\orcidID{0000-0002-1355-7663}%
}
\authorrunning{D. Magiera et al.}
\institute{%
Centre for Cognitive Science \& Institute of Psychology, Technische Universität Darmstadt, Germany%
 \and
Honda Research Institute Europe GmbH, Offenbach am Main, Germany%
}
\maketitle              %

\begin{abstract}
Humans can often acquire and synthesize complex, recursive concepts from minimal experience. Leveraging cognitive
insights, we propose the Minimalist Machine, a framework for inductive program synthesis designed to model such
conceptual learning. The system uses a compact relational subset of Prolog: Programs are searched within a fixed schema
of body-free facts and two-body conjunctive Horn clauses. Recursion is not defined by a dedicated metarule. Instead, it
emerges when a target predicate is reused inside the body of a learned clause. Inspired by a primary school curriculum,
the model is taught through a human-curated, sequential introduction of new concepts in arithmetic. Starting from
initially empty knowledge base, it first acquires simple structural predicates, then successor-based state
transformations, and finally recursive programs for addition, subtraction, multiplication, and division. Ultimately,
this approach yields the fully transparent, inductive reasoning trace necessary for human-like conceptual learning.

\emph{The full code and curriculum are available online at
\href{https://github.com/cognitive-modeling/minimalist-machine/tree/IJCLR-2026}{github.com/cognitive-modeling/minimalist-machine}.}

\keywords{Inductive Program Synthesis \and Curriculum Learning \and Recursive Programs}
\end{abstract}

\section{Introduction}\label{introduction}

Human learning is open-ended. Humans continually revise and extend their internal models through observation,
interaction, and exploration rather than optimizing a fixed objective over a fixed dataset \cite{Dupoux_etal_2026}. This
remains difficult for current AI systems, which are typically trained on fixed datasets and often require human-guided
retraining or fine-tuning to adapt to new environments \cite{Hadsell_etal_2020}. Large language models also continue to
struggle with structural reasoning tasks that require systematic abstraction, such as Bongard visual puzzles
\cite{Bongard_1970,Depeweg_etal_2024,Wuest_etal_2025}. This raises a central question: how can a machine acquire new
structured concepts in a cumulative, reusable, and open-ended way?

A natural formal setting for this problem is Inductive Programming (IP), which studies methods for learning computer
programs from incomplete specifications, such as input-output examples \cite{Flener_Schmid_2008}. This perspective is
attractive because programs can represent procedures, abstractions, and reusable forms of reasoning. A central mechanism
for making such learning cumulative is predicate invention. By inventing new predicates, a learner can extend its own
knowledge base with reusable abstractions, allowing later programs to build on concepts learned in earlier tasks
\cite{Muggleton_etal_2018,Rios_Matwin_1998}. This makes program induction more than the search for a single
task-specific solution; it becomes a process of constructing a growing library of concepts. A core target for such
cumulative program induction is recursion. Many cognitive abilities, including language, planning, and arithmetic,
depend on applying the same operation repeatedly while maintaining a structured representation of intermediate states.
In arithmetic, for example, children do not memorize every possible sum or product. They learn procedures such as
counting, incrementing, repeated addition, and grouping. Previous work has shown that recursive equations can be induced
from examples by detecting structural regularities \cite{Kitzelmann_2009}. The harder problem is to discover recursive
structure in a way that supports reuse across tasks, rather than solving each task in isolation.

Existing approaches to Inductive Programming and Inductive Logic Programming make this problem tractable by imposing
strong assumptions about the hypothesis space. Systems such as \emph{Metagol} and \emph{Igor2} show that programs can be
inferred from examples and can generalize compositionally \cite{Cropper_Muggleton_2016,Kitzelmann_2009}. However, these
systems typically rely on predefined languages, metarules, functional schemes, mode declarations, type declarations,
predicate declarations, or other structural restrictions that depend on the target concepts and determine in advance
which kinds of programs can be found \cite{Cropper_Dumancic_2020}. These forms of language bias are necessary to control
the search space, but they also limit what the learner can invent. The open question is therefore what is a minimal
structure that must be built into the learner before it can discover reusable recursive programs.

\subsubsection{Novelty and Contributions.}\label{novelty-and-contributions.}

We ask whether a machine can learn arithmetic in a child-like way, starting from simple relational tasks and progressing
through a curriculum toward recursive arithmetic. Our main contributions are:

\begin{enumerate}
\def\labelenumi{\arabic{enumi}.}
\item
  \emph{Discovery of recursive programs with a minimal language bias.} We show that recursive symbolic programs can be
  discovered using a small set of schemas for relational instances and two-body Horn clauses with only three unique
  variables. The system is not given dedicated templates for recursion, arithmetic operations, accumulation, or inverse
  relations.
\item
  \emph{Teaching arithmetic from an initially empty knowledge base.} We show that a machine starting from an empty
  program can acquire arithmetic through a human-inspired curriculum. It first learns simple relational concepts such as
  projection, equality, permutation, and successor-based state transformation. These concepts are then used as
  primitives in recursive programs for addition, subtraction, multiplication, and division.
\end{enumerate}

\section{The Minimalist Machine}\label{the-minimalist-machine}

The Minimalist Machine follows the idea that complex structure can arise from a small set of general operations. Our
goal is to reduce the machine's assumptions to a minimal set while preserving its ability to discover complex symbolic
programs like multiplication. Instead of supplying task-specific templates, the machine searches over a fixed schema of
small relational clauses. Its operations are limited to introducing simple facts, constructing pair-structured terms,
connecting predicates through shared variables, and composing two known relations by conjunction. Therefore, recursion
is not provided by a dedicated meta-rule. It emerges when the target predicate is selected for reuse inside the body of
a clause. We decided to make the machine relational rather than functional. Its learned programs can therefore be viewed
as a restricted subset of Prolog, consisting of simple Horn clauses over pair-structured terms. A function specifies how
to compute an output from given inputs. A relation specifies a constraint among values and can therefore be used in
multiple directions: to compute an output, infer a missing input, check consistency, or express inverse structure. This
makes relational programs suitable for cumulative program induction, because each learned predicate can become a
reusable abstraction for later tasks.

Program induction is therefore not treated as the solution of isolated tasks. It is treated as the construction of a
growing library of concepts. Each learned term becomes a reusable cognitive primitive that can be recombined and
extended. Learning is cumulative: projection predicates support equality predicates, equality predicates support
permutations, permutations support state transformations, and these transformations support recursive arithmetic
programs.

\subsection{Minimal Language Bias}\label{minimal-language-bias}

Most inductive logic programming systems control learning through a predefined language bias. Systems such as
\emph{Popper}, \emph{Aleph}, and \emph{Metagol} allow the user to specify which predicates may be used, how clauses may
be shaped, and which forms of recursion or invention are permitted \cite{Cropper_Dumancic_2020}. Such restrictions are
necessary because the unrestricted space of possible logic programs is too large to search effectively. The Minimalist
Machine uses only a fixed relational schema as its language bias. Candidate programs are generated from two kinds of
clauses: body-free facts and two-body Horn rules. Instances introduce symbols, pair-structured terms, and
projection-like relations. Rules compose two known predicates by wiring their arguments through a small set of template
variables. The target predicate itself may appear in the body, allowing recursion to arise as an ordinary case of
relational composition rather than through a dedicated recursive meta rule.

\begin{table}[t]
\centering
\small
\begin{tabular}{l|l}
\hline
\textbf{Type} & \textbf{Schema} \\
\hline
Fact
    &   $p.$                 \\
    &   $p((X,Y)).$        \\
    &   $p(((X,Y),Y)).$   \\
    &   $p(((Y,X),Y)).$   \\
Rule
    & $p((X,Y)) \mathrel{\texttt{:-}} p_0((V_0,V_1)),\; p_1((V_2,V_3)).$ \\
\hline
\end{tabular}
\vspace{0.8em}
\caption{\textbf{Minimal set of candidate clause schemas.}
For facts, $X$ and $Y$ may remain variables or be replaced by known symbols from $\mathcal{B}$.
For rules, $V_0, V_1, V_2,V_3$ denote body-argument positions, each ranging over $\{X,Y,Z\}$.
The body predicates satisfy $p_0,p_1\in\mathcal{B}\cup\{p\}$, where $\mathcal{B}$ is the current set of known unary predicates available as background knowledge.}
\label{tab:clause-grammar}
\end{table}

The machine searches over a restricted Horn-clause hypothesis space. Each learned target predicate \(p\) is represented
as an unary predicate over pair-structured terms. This encoding allows binary relations to be represented as unary
predicates whose single argument is a pair. For example, a relation between \(X\) and \(Y\) is represented as
\(p((X,Y))\). Here, \((X,Y)\) is syntactic shorthand for the pair term \(\mathrm{and}(X,Y)\), reflecting the system's
use of the conjunction operator to construct structured terms.\footnote{In Prolog, this is commonly implemented using
  the comma functor, written as \texttt{\textquotesingle{},\textquotesingle{}(X,Y)} or, in infix notation,
  \texttt{(X,Y)}.}

Let \(\mathcal{V}=\{X,Y,Z\}\) be the set of template variables and \(\mathcal{B}\) be the current background knowledge,
including a set of known symbols. These symbols are not separated into distinct sets for constants and predicates. In
fact templates, a symbol from \(\mathcal{B}\) may be inserted as an argument-free constant. In rule templates, a symbol
from \(\mathcal{B}\) may be selected as a predicate. Thus, the same learned symbol can participate both as a value in
structured terms and as a predicate available for later relational composition.

The fact language contains one nullary fact and three unary fact schemas:
\[ p, \qquad p((X,Y)), \qquad p(((X,Y),Y)), \qquad p(((Y,X),Y)). \] In the unary fact schemas, \(X\) and \(Y\) may
either remain variables or be replaced by known symbols from \(\mathcal{B}\), where those symbols are used as
argument-free constants. The first form introduces a nullary fact when the target has arity \(0\). The remaining forms
introduce pair facts and projection-like relations over pair-structured terms. The rule language has the fixed form
\[p((X,Y)) \mathrel{\texttt{:-}} p_0((V_0,V_1)),\; p_1((V_2,V_3)),\] where \(p_0,p_1\in\mathcal{B}\cup\{p\}.\) The
symbols \(V_0,\ldots,V_3\) denote body-argument positions, not additional variables. Each \(V_i\) is instantiated by one
of the three template variables \(X\), \(Y\), or \(Z\):
\[ V_i \in \mathcal{V}=\{X,Y,Z\} \quad \mathrm{for} \quad i\in\{0,1,2,3\}. \] Each body predicate receives an ordered
pair of variables drawn from \(\mathcal{V}\times\mathcal{V}\). Thus, body predicates may use repeated-variable patterns,
such as \((X,X)\), or cross-variable patterns, such as \((X,Z)\). These wirings allow the schema to express shared-value
joins, relational composition, and variable reordering while using only three unique variables. The full hypothesis
space for a target predicate is \[ \mathcal{H}=\mathcal{F}\cup\mathcal{R}, \] where \(\mathcal{F}\) is the set of
candidate facts and \(\mathcal{R}\) is the set of candidate rules. Allowing \(p\) itself as a possible body predicate
makes recursion a special case of the same rule schema. For example, a repeated-step relation can be represented as
\[p((X,Y)) \mathrel{\texttt{:-}} q((X,Z)), p((Z,Y)).\] In this way, recursive programs are discovered as ordinary
members of the same relational search space, rather than being introduced through a special recursive rule template.
This yields a compact hypothesis language. Learning is constrained to a small set of clause types: facts, projections,
relational intersections, shared-value joins, variable rewirings, and relational composition. Recursive chaining is
handled as a special case of composition when the target predicate appears in the body.

The Minimalist Machine learns a new target predicate from a finite set of examples \(E\). Positive examples are written
as ground atoms, while negative examples are written using \texttt{not}. For a target predicate such as \texttt{add},
which relates a pair of numbers to their sum, the example set may be written as \[
E =\left\{
\begin{array}{l}
\mathrm{add}((1,1),2),\\
\mathrm{add}((2,1),3),\\
\mathrm{add}((2,2),4),\\
\mathrm{not}(\mathrm{add}((1,2),4)),\\
\mathrm{not}(\mathrm{add}((2,1),4))
\end{array}
\right\}.
\] The machine searches the all candidate clauses from \(\mathcal{H}\) and tests each candidate against \(E\). A
candidate hypothesis \(h\) is accepted if it returns the correct result for every query in \(E\). Learning a predicate
is therefore a constrained search problem: the schemas define the possible clauses, and the labelled examples constrain
the search toward clauses that solve the intended relation.

\subsection{Programs as Concept Libraries}\label{programs-as-concept-libraries}

The Minimalist Machine treats learned programs as a growing library of learned predicates. Once a predicate has been
learned, its symbol becomes available in the background knowledge \(\mathcal{B}\) for later tasks. This allows the
machine to construct increasingly complex programs from previously induced structure. For example, if the machine has
already learned predicates such as \(\mathrm{parent}\) and \(\mathrm{male}\), it can later define \(\mathrm{father}\) by
composing them: a father is someone who is both a parent and male. This cumulative structure is also important for
interpretability. Each learned program is expressed in terms of previously learned predicates, so the resulting
definitions provide an explicit inductive trace. Rather than producing a single opaque solution for each task, the
system builds a sequence of reusable abstractions. This connects program induction with the broader goal of constructing
generalizable and interpretable knowledge \cite{Rule_etal_2020}.

\section{Teaching Arithmetic}\label{teaching-arithmetic}

The assumptions of the Minimalist Machine are intentionally small, as described in the previous chapter. The machine
generates only simple facts and two-body Horn rules over pair-structured terms. This compact language bias makes the
search space finite, but it does not by itself make open-ended learning possible. Even a finite hypothesis space can be
too large to search effectively. Especially when the learner is asked to discover a complex mathematical concept before
the necessary intermediate predicates have been learned.

\begin{figure}
\centering
\includegraphics[width=3.25in]{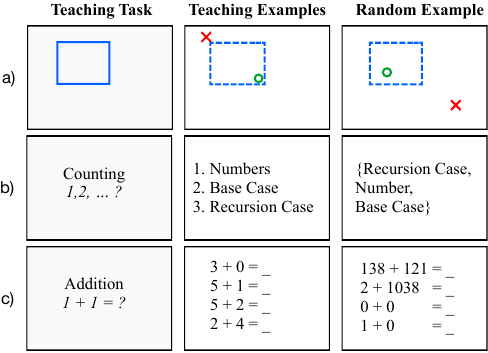}
\caption{(a) Possible rectangle-game scenarios from the teaching game \cite{Shafto_Goodman_2008}. (b) Order of learned
concepts for counting. (c) Examples for learning addition.}\label{fig:teaching}
\end{figure}

A second constraint is therefore needed: teaching. This is illustrated by pedagogical learning tasks such as the
rectangle game \cite{Shafto_Goodman_2008}. In the rectangle game, a teacher tries to communicate a rectangle concept by
selecting informative positive and negative examples. Importantly, in this game the examples are not random. Teachers
tend to choose examples that are useful for the learner, such as positive examples near corners and negative examples
near boundaries. As the students know that the examples are not random and assume the teacher is helpful, they can learn
the intended rectangle more quickly than from randomly sampled data (see Fig.~\ref{fig:teaching}). In realistic teaching
scenarios, e.g.~in schools, learning depends not only on which examples are shown, but also on why those examples are
shown and in what order. Positive and negative examples teach important cases of a target relation, while the curriculum
determines which target relation is learned next. A teacher therefore constrains learning in two ways: by selecting
examples that distinguish the intended predicate, and by ordering tasks so that simpler predicates become available
before more complex ones are introduced. This view also changes the role of examples. Examples are not only evidence for
a single target predicate. They also act as teaching signals that decide which predicate is added next to the background
knowledge \(\mathcal{B}\). Once a predicate has been learned, its symbol becomes available for later facts and rules.
The curriculum therefore turns program induction into a cumulative process: earlier predicates reduce the search space
for later predicates by providing reusable structure.

\subsection{Human-Inspired Curriculum}\label{human-inspired-curriculum}

\begin{figure}
\centering
\includegraphics[width=4.75in]{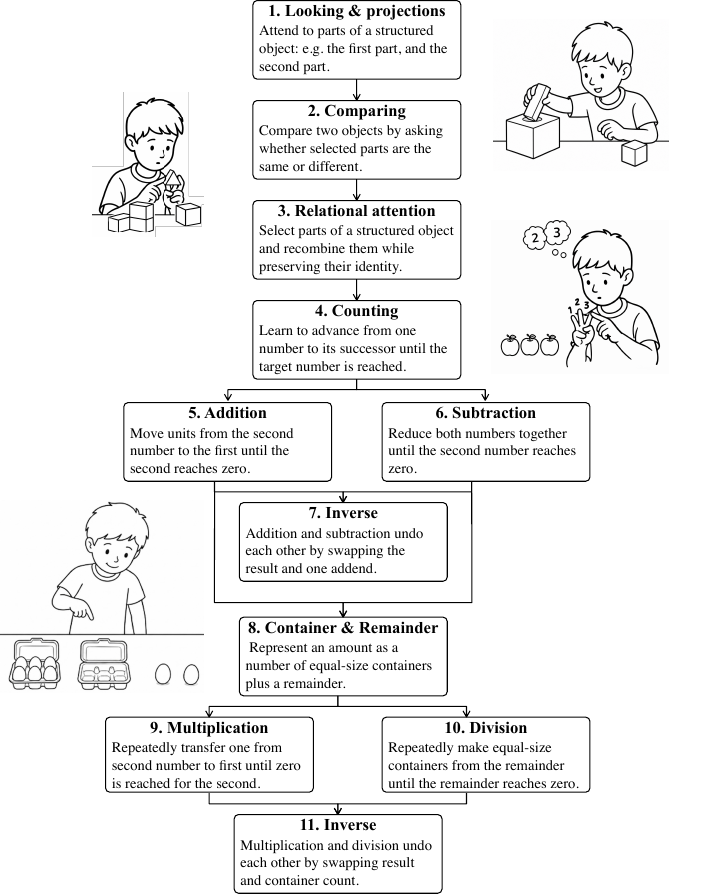}
\caption{\textbf{A human-inspired curriculum for basic arithmetic.} The curriculum begins with structural fundamentals
(1--3), then introduces counting and state-based addition and subtraction (4--7), and finally extends the same
relational structure to container-and-remainder reasoning for multiplication and division (8--11). \emph{Child images
created with GPT-5.5.}}\label{fig:curriculum}
\end{figure}

In this paper, we want to teach the Minimalist Machine basic arithmetic. In order to do so, we structure the learning
process along the lines of primary school curricula. Fig.~\ref{fig:curriculum} shows an overview of our curriculum. It
is organized as a sequence of increasingly complex relational tasks. The learning machine first acquires symbols for
numbers. It then learns successor relations, which define the order of the number sequence. These relations support
state-transition predicates such as transferring one unit from one part of a state to another. Finally, these state
transitions are combined with base cases to form recursive arithmetic programs. This approach is related to pedagogical
models of learning, where examples are selected by a teacher to guide the learner toward a concept
\cite{Shafto_Goodman_2008,Shafto_etal_2014}. It is also related to work on modular concept discovery and conceptual
bootstrapping, where previously learned concepts make later learning problems tractable
\cite{Dechter_etal_2013,Hocquette_etal_2024,Zhao_etal_2023}. In the Minimalist Machine, this idea is implemented
directly: the learned predicates form a growing knowledge base, and the curriculum determines the order in which this
background knowledge is constructed.

\subsubsection{Fundamentals: Looking, Comparing, and Relational
Attention.}\label{fundamentals-looking-comparing-and-relational-attention.}

Before the machine can learn arithmetic, it first has to learn how to inspect and reorganize pair-structured structures.
These early predicates form the structural foundation of the curriculum. They do not perform arithmetic themselves.
Instead, they make later arithmetic predicates possible by giving the learner ways to look at parts of a state, compare
parts across structures, and move attention between positions. Therefore, the first step is looking. Since the system
represents structured data using binary pair terms, the learner must acquire predicates that access positions inside
such terms. The machine first learns the predicates first part and second part, which select the two immediate
components of a pair. Nested projections such as \texttt{first\_of\_first} and \texttt{second\_of\_first} then access
positions inside the first component. For example, if an argument has the form \texttt{((a,b),c)}, then
\texttt{first\_of\_first} refers to \texttt{a} and \texttt{second\_of\_first} refers to \texttt{b}. The second step is
comparing. After the learner can attend to positions, it can learn predicates that test whether two structures share a
part. For example, \texttt{equal\_first} can test whether \texttt{(a,b)} and \texttt{(a,c)} share the same first
component, while \texttt{equal\_second} can test whether \texttt{(b,a)} and \texttt{(c,a)} share the same second
component. Cross-position predicates such as \texttt{equal\_first\_first\_to\_first\_second} compare different positions
across two terms. For example, this predicate can test whether \texttt{((a,b),c)} and \texttt{((d,a),e)} share the value
\texttt{a} in different nested positions. These predicates allow the machine to express sameness without encoding a
whole nested pattern directly in the clause head. The third step is relational attention. Once positions can be accessed
and compared, the learner can construct predicates that rearrange attention over a state without changing the underlying
values. Mapping and permutation predicates, such as \texttt{map\_inner\_first\_and\_second\_to\_first} or
\texttt{swap\_inner\_first\_and\_second}, describe how selected parts of one structured term correspond to selected
parts of another. \texttt{Swap\_inner\_first\_and\_second} can relate \texttt{((a,b),c)} to \texttt{((c,b),a)} by
swapping the inner first position with the outer second position. This makes it possible to express transformations
through local positional relations, which can later be reused to define inverse mappings between arithmetic operations.

These three steps prepare the learner for arithmetic. Looking provides access to structured input, comparing identifies
which parts are preserved and relational attention describes how parts are rearranged or carried forward.

\subsubsection{Counting, Addition, and Subtraction.}\label{counting-addition-and-subtraction.}

Counting is the first arithmetic level of the curriculum. The learner first acquires number symbols such as
\texttt{zero}, \texttt{one}, and \texttt{two}. These symbols are then taught by the successor predicate
\texttt{next\ num}, which relates each number to its next number in the counting sequence. This gives the machine an
ordered number line without giving it arithmetic operations directly. Additional helper predicates, such as
\texttt{next\_first} and \texttt{next\_second}, apply this successor relation to selected positions inside a
pair-structured state. Counting is therefore learned as a relational structure over symbols: the machine learns which
symbol follows which other symbol, and later predicates can reuse this ordering to express numerical change.

After that, addition is introduced through the state-transition predicate \texttt{transfer\_one}. A state such as
\texttt{(two,\ three)} represents two numbers. The predicate \texttt{transfer\_one} relates this pair to
\texttt{(three,\ two)}, transferring one unit from the second number to the first while preserving the total. In Prolog
notation, this is defined by composing two learned successor-based mappings: \[
\vbox{
\hbox{\verb|transfer_one((X, Y)) :-|}
\hbox{\verb|    map_next_first((X, Y)),|}
\hbox{\verb|    map_next_second((Y, X)).|}
}
\] \noindent Here, \texttt{map\_next\_first} increases the first number using the learned successor relation, while
\texttt{map\_next\_second} is used in reverse to decrease the second number. For example,
\texttt{?-\ transfer\_one(((two,\ three),\ Result)).} can return \texttt{Result\ =\ (three,\ two)}. Addition is then
learned by combining this repeated transition with a base case. The base case states that when the second quantity is
\texttt{zero}, the first quantity is the result. The recursive case repeatedly applies \texttt{transfer\_one} until this
base case is reached. Thus, addition is not learned as a table of sums, but as a recursive relation over state changes.

Subtraction is similarly learned through the related transition predicate \texttt{reduce\_both}. Instead of moving one
unit from one quantity to another, \texttt{reduce\_both} decrements both the minuend and the amount to be subtracted. A
state such as \texttt{(five,\ three)} is related to \texttt{(four,\ two)}, then to \texttt{(three,\ one)}, and finally
to \texttt{(two,\ zero)}. As in addition, the base case is reached when the second quantity is zero. The first quantity
is then the result. Subtraction is therefore learned as recursive countdown over a pair-structured state. The relation
reuses the same successor-based knowledge as addition, but applies it to a different state transformation. The inverse
relation between addition and subtraction is then taught by showing that the same structured positions can be mapped
differently: an addition state \texttt{((a,b),c)} corresponds to a subtraction state \texttt{((c,b),a)}.

\subsubsection{Container and Remainder.}\label{container-and-remainder.}

The next arithmetic level introduces a container-and-remainder representation. A number can be represented as a
structured state consisting of a number of equal-sized containers and a remainder amount. For example, the state
\texttt{((three,\ two),\ zero)} represents three containers of size two with zero remainder. This representation
prepares the learner for multiplication and division. Instead of treating multiplication as a primitive operation, the
machine first learns how to represent a total amount as containers plus a remainder. The predicate
\texttt{start\_with\_zero\_remainder} introduces this representation by pairing an input container state with a zero
remainder.

Regrouping a container into the remainder amount is the state transition used for multiplication. In a state such as
\texttt{((one,\ ten),\ two)}, there is one container of size ten and a remainder of two. Opening the container removes
one container and adds its size to the remainder, producing a state with \texttt{zero} containers and \texttt{twelve} as
the remainder. The predicate \texttt{add\_container\_size\_to\_remainder} captures the part of this transformation that
adds the container size to the remainder while preserving the container structure. This allows multiplication to be
learned as repeated opening of containers: each recursive step moves one container-size amount into the remainder.

Making a container from the remainder amount is the inverse state transition used for division. Here, the state is
represented as \texttt{((Remainder,\ ContainerSize),\ Count)}, where the first component stores the remaining amount The
second stores the container size, and the outer second component counts how many containers have been made. For example,
((six, two), zero) can transition to ((four, two), one), then to \texttt{((two,\ two),\ two)}, and finally to ((zero,
two), three). The predicate \texttt{take\_container\_size\_from\_remainder} captures the step that subtracts one
container-size amount from the remainder while preserving the container size. Together with a counter update, this lets
division be learned as repeatedly making equal-sized containers until no remainder is left.

\subsubsection{Multiplication and Division.}\label{multiplication-and-division.}

\begin{figure}
\centering
\includegraphics[width=4.75in]{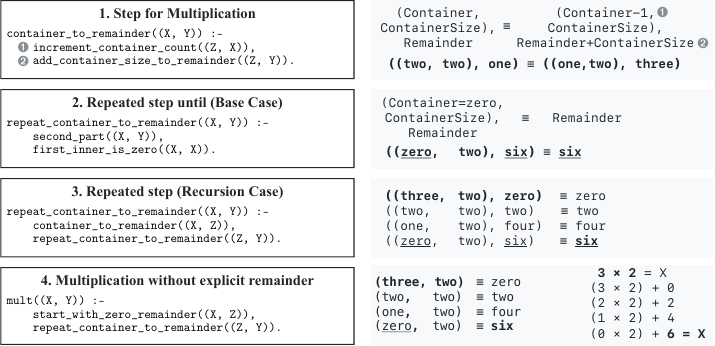}
\caption{\textbf{Learning multiplication as recursive container regrouping.} The learned program opens one container at
a time, adds its size to the remainder, and stops when the container count reaches zero. The advance predicate
\protect\texttt{mult} hides the explicit remainder state while preserving the learned recursive
structure.}\label{fig:results-multiplication}
\end{figure}

Multiplication is taught as repeated regrouping from containers into a remainder amount as shown in
Fig.~\ref{fig:results-multiplication}. A multiplication problem such as \texttt{mult(((three,\ two),\ Result))} is first
converted into the state \texttt{((three,\ two),\ zero)}, meaning three containers of size two and zero remainder. Each
recursive step opens one container: the container count is reduced by one, while the container size is added to the
remainder. Thus, the internal state sequence for \(3 \times 2\) is \texttt{((three,\ two),\ zero)},
\texttt{((two,\ two),\ two)}, \texttt{((one,\ two),\ four)}, and \texttt{((zero,\ two),\ six)}. The base case is reached
when the container count is \texttt{zero}; the remainder is then returned as the result. Multiplication is therefore not
introduced as a dedicated arithmetic template. It is discovered as recursive state change over a container-and-remainder
representation.

Division is learned as the inverse process: repeatedly making containers from a remainder amount. A division problem
such as \texttt{div(((six,\ two),\ Result))} is converted into the state \texttt{((six,\ two),\ zero)}, meaning six
units of remainder, container size two, and zero containers made. Each recursive step removes one container-size amount
from the remainder and increases the count of made containers by one. The internal state sequence is
\texttt{((six,\ two),zero)}, \texttt{((four,\ two),one)}, \texttt{((two,\ two),\ two)}, and
\texttt{((zero,\ two),\ three)}. The base case is reached when the remainder is zero and the number of made containers
is then returned as the result. This version models exact division at the level of early elementary arithmetic.

\section{Results}\label{results}

We evaluated the Minimalist Machine by teaching the full arithmetic curriculum from examples (positive and negative).
The learner was initialized with an empty knowledge base \(\mathcal{B}\), and tasks were presented sequentially in the
curriculum order. After each task, the first candidate clause that correctly solved the examples was asserted into
long-term memory and became part of the background knowledge \(\mathcal{B}\) for all later tasks. For each target
predicate, the machine searched the same fixed candidate space described earlier. Candidate clauses were generated in a
deterministic order: facts were tried first, followed by two-body Horn rules. Each candidate was tested against the
labelled examples for the current task. Positive examples had to succeed, and examples marked with not had to fail.
Search stopped when a candidate solved all examples or when the task-specific iteration limit was reached. We used a
custom Warren Abstract Machine \cite{Warren_1983} compiler implemented in Python. The learned Horn clauses should behave
similarly in standard Prolog systems such as SWI-Prolog.

\begin{figure}
\centering
\includegraphics[width=4.75in]{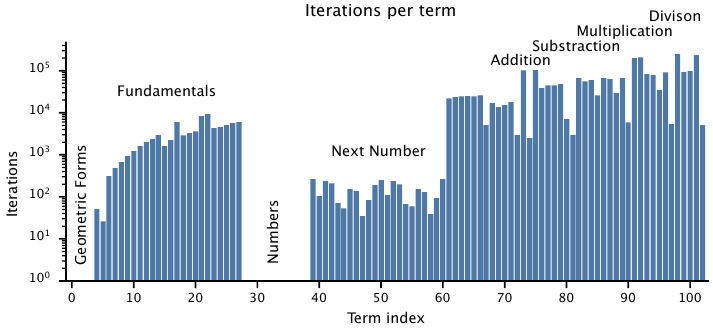}
\caption{\textbf{Iterations required for discovering the correct rule.} The number of candidate clauses tested before
the machine finds a rule consistent with the examples at each curriculum step.}\label{fig:results-iterations}
\end{figure}

Fig. 4 shows that search effort increases with the complexity of the learned predicates. Early tasks such as learning
the number symbols are solved almost immediately, because they are represented by simple facts. The structural
fundamentals require more iterations, because the learner must discover projection, comparison, mapping, and permutation
rules. Successor and number predicates remain comparatively cheap, mostly in the range of tens to hundreds of
iterations. The largest increases appear once the curriculum reaches recursive arithmetic: the first recursive rule is
discovered after 73 terms, where the learner finds the recursive case for addition after learning its base case.
Subtraction, multiplication, and division then require longer searches, with several later tasks reaching around
\(10^5\) iterations. Overall, later arithmetic predicates remain reachable, but they are harder to discover because the
growing number of learned symbols in \(\mathcal{B}\) increases the hypothesis space for every new predicate.

\begin{figure}
\centering
\includegraphics[width=4.75in]{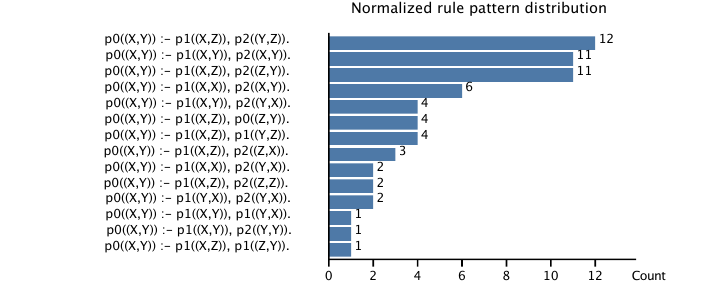}
\caption{\textbf{Learned normalized rule patterns.} The learned rule predicates are shown in a shared relational format,
illustrating how the same two-body Horn-clause schema expresses projections, state transitions, recursion, and inverse
mappings.}\label{fig:results-rules}
\end{figure}

After normalization, the most frequent rules correspond to a small set of dyadic clause patterns familiar from
meta-interpretive learning \cite{Muggleton_etal_2015} as shown in Figure 5. Many learned clauses instantiate a
\emph{chain} pattern, such as \emph{p((X, Y)):- p0((X, Z)), p1((Z, Y))}, where an intermediate variable links two
relations. Other clauses express \emph{shared-variable joins}, such as \emph{p((X, Y)):- p0((X, Z)), p1((Y, Z))}, where
two arguments are constrained to share the same value. Recursive arithmetic clauses instantiate a \emph{tail-recursive}
pattern, such as \emph{p((X, Y)):- p0((X, Z)), p((Z, Y))}, where the target predicate reappears in the body.

\subsubsection{Bidirectional Extrapolation.}\label{bidirectional-extrapolation.}

Because the learned arithmetic predicates are relational, they can be queried in more than one direction -- as they
should be in Prolog. For example, multiplication can be used not only to compute a result from two given factors, but
also to search for factor pairs that produce a given result. To ask which factors multiply to six, the factors can be
left as variables: \texttt{?-\ mult(((X,\ Y),\ six)).} and is answered with:

\begin{verbatim}
X = one,   Y = six ;
X = two,   Y = three ;
X = three, Y = two ;
X = six,   Y = one.
\end{verbatim}

The exact set of answers depends on the available number symbols and successor relations in the current background
knowledge \(\mathcal{B}\).

\section{Discussion}\label{discussion}

The results show that recursive arithmetic programs can be discovered from a compact relational search space when
learning is organized as a curriculum. The Minimalist Machine is not given dedicated templates. Instead, these programs
are assembled from a small set of reusable relational forms. The curriculum is essential for making this possible.
Without previously learned predicates, the search space for later arithmetic tasks would be too large and poorly
structured. The results show that later predicates become harder to discover as the background knowledge \(\mathcal{B}\)
grows, because more learned symbols increase the candidate space. At the same time, this growing background knowledge is
what makes later concepts expressible. The curriculum therefore plays a dual role: it increases the space of possible
clauses, but also supplies the intermediate abstractions needed to solve more complex tasks. This tradeoff is central to
cumulative program induction. Designing an effective curriculum is, of course, non-trivial and shifts a lot of the hard
work from the learner to the teacher. It may well be that the reason why children can learn arithmetic at all is that
their teachers provide a lot of scaffolding.\\
The teacher also has to provide positive and negative examples for each predicate in the curriculum. Future work should
investigate how the learner can ask diagnostic questions to generate examples more efficiently in interaction with a
teacher.

The work also relates to lifelong and continual learning. Continual learning is often difficult for current
machine-learning systems because new tasks can interfere with previously learned knowledge, producing catastrophic
forgetting \cite{Ratcliff_1990,Mundt_etal_2020,Ven_etal_2025}. The Minimalist Machine uses a different mechanism.
Learned predicates are not overwritten by gradient updates; they are added to an explicit background knowledge base.
This makes reuse direct and avoids forgetting in the narrow sense that earlier learned rules remain available. However,
this also creates a different problem: the background knowledge grows, and the search space grows with it. Future work
must therefore study how a learner can organize and revise its learned predicates as the concept library expands.

Overall, the results suggest that recursive arithmetic can emerge from composing two known predicates, given a
curriculum that introduces useful intermediate predicates in a teachable order. The Minimalist Machine therefore shows
how simple symbolic relations can be organized into recursive arithmetic programs.

\begin{credits}
\subsubsection{\ackname}
This work was supported by the Honda Research Institute Europe (HRI-EU).

\subsubsection{\discintname}
The authors have no competing interests to declare that are relevant to the content of this article.

\end{credits}
\bibliographystyle{splncs04_unsrt}
\bibliography{references}

@misc{Cropper_Dumancic_2020,
  author = {Cropper, Andrew and Dumančić, Sebastijan},
  publisher = {arXiv},
  title = {Inductive Logic Programming at 30: {{A}} New Introduction},
  year = {2020},
  doi = {10.48550/ARXIV.2008.07912}
}

@inproceedings{Dechter_etal_2013,
  author = {Dechter, Eyal and Malmaud, Jonathan and Adams, Ryan P. and
    Tenenbaum, Joshua B.},
  title = {Bootstrap Learning via Modular Concept Discovery},
  booktitle = {Proceedings of the 23rd international joint conference on
    artificial intelligence},
  pages = {1302-1309},
  year = {2013}
}

@inbook{Rios_Matwin_1998,
  author = {Rios, Riverson and Matwin, Stan},
  publisher = {Springer Berlin Heidelberg},
  title = {Predicate Invention from a Few Examples},
  booktitle = {Advances in artificial intelligence},
  pages = {455-466},
  year = {1998},
  doi = {10.1007/3-540-64575-6_70}
}

@misc{Hocquette_etal_2024,
  author = {Hocquette, Céline and Langer, Johannes and Cropper, Andrew
    and Schmid, Ute},
  publisher = {arXiv},
  title = {Can Humans Teach Machines to Code?},
  year = {2024},
  doi = {10.48550/ARXIV.2404.19397}
}

@article{Muggleton_etal_2018,
  author = {Muggleton, Stephen H. and Schmid, Ute and Zeller, Christina
    and Tamaddoni-Nezhad, Alireza and Besold, Tarek},
  publisher = {Springer Science; Business Media LLC},
  title = {Ultra-Strong Machine Learning: {{Comprehensibility}} of
    Programs Learned with {{ILP}}},
  journal = {Machine Learning},
  volume = {107},
  number = {7},
  pages = {1119-1140},
  year = {2018},
  month = {may},
  doi = {10.1007/s10994-018-5707-3}
}

@article{Hadsell_etal_2020,
  author = {Hadsell, Raia and Rao, Dushyant and Rusu, Andrei A. and
    Pascanu, Razvan},
  publisher = {Elsevier BV},
  title = {Embracing Change: {{Continual}} Learning in Deep Neural
    Networks},
  journal = {Trends in Cognitive Sciences},
  volume = {24},
  number = {12},
  pages = {1028-1040},
  year = {2020},
  month = {dec},
  doi = {10.1016/j.tics.2020.09.004}
}

@inbook{Kitzelmann_2009,
  author = {Kitzelmann, Emanuel},
  publisher = {Springer Berlin Heidelberg},
  title = {Analytical Inductive Functional Programming},
  booktitle = {Logic-based program synthesis and transformation},
  pages = {87-102},
  year = {2009},
  doi = {10.1007/978-3-642-00515-2_7}
}

@inbook{Ven_etal_2025,
  author = {Ven, Gido M. van de and Soures, Nicholas and Kudithipudi,
    Dhireesha},
  publisher = {Elsevier},
  title = {Continual Learning and Catastrophic Forgetting},
  booktitle = {Learning and memory: A comprehensive reference},
  pages = {153-168},
  year = {2025},
  doi = {10.1016/b978-0-443-15754-7.00073-0}
}

@article{Ratcliff_1990,
  author = {Ratcliff, Roger},
  title = {Connectionist Models of Recognition Memory: {{Constraints}}
    Imposed by Learning and Forgetting Functions},
  journal = {Psychological Review},
  volume = {97},
  number = {2},
  pages = {285-308},
  year = {1990},
  doi = {10.1037/0033-295X.97.2.285}
}

@misc{Dupoux_etal_2026,
  author = {Dupoux, Emmanuel and LeCun, Yann and Malik, Jitendra},
  publisher = {arXiv},
  title = {Why {{AI}} Systems Don’t Learn and What to Do about It:
    {{Lessons}} on Autonomous Learning from Cognitive Science},
  year = {2026},
  doi = {10.48550/ARXIV.2603.15381}
}

@misc{Mundt_etal_2020,
  author = {Mundt, Martin and Hong, Yongwon and Pliushch, Iuliia and
    Ramesh, Visvanathan},
  publisher = {arXiv},
  title = {A Wholistic View of Continual Learning with Deep Neural
    Networks: {{Forgotten}} Lessons and the Bridge to Active and Open
    World Learning},
  year = {2020},
  doi = {10.48550/ARXIV.2009.01797}
}

@misc{Cropper_Muggleton_2016,
  author = {Cropper, Andrew and Muggleton, Stephen H.},
  publisher = {https://github.com/metagol/metagol},
  title = {Metagol System},
  year = {2016}
}

@article{Muggleton_etal_2015,
  author = {Muggleton, Stephen H. and Lin, Dianhuan and
    Tamaddoni-Nezhad, Alireza},
  publisher = {Springer Science; Business Media LLC},
  title = {Meta-Interpretive Learning of Higher-Order Dyadic Datalog:
    {{Predicate}} Invention Revisited},
  journal = {Machine Learning},
  volume = {100},
  number = {1},
  pages = {49-73},
  year = {2015},
  month = {mar},
  doi = {10.1007/s10994-014-5471-y}
}

@article{Rule_etal_2020,
  author = {Rule, Joshua S. and Tenenbaum, Joshua B. and Piantadosi,
    Steven T.},
  publisher = {Elsevier BV},
  title = {The Child as Hacker},
  journal = {Trends in Cognitive Sciences},
  volume = {24},
  number = {11},
  pages = {900-915},
  year = {2020},
  month = {nov},
  doi = {10.1016/j.tics.2020.07.005}
}

@techreport{Warren_1983,
  author = {Warren, David H. D.},
  publisher = {Artificial Intelligence Center, SRI International},
  title = {An Abstract Prolog Instruction Set},
  year = {1983},
  month = {oct},
  address = {Menlo Park, CA, USA},
  note = {Archived PDF, original published October 1983}
}

@article{Zhao_etal_2023,
  author = {Zhao, Bonan and Lucas, Christopher G. and Bramley, Neil R.},
  publisher = {Springer Science; Business Media LLC},
  title = {A Model of Conceptual Bootstrapping in Human Cognition},
  journal = {Nature Human Behaviour},
  volume = {8},
  number = {1},
  pages = {125-136},
  year = {2023},
  month = {oct},
  doi = {10.1038/s41562-023-01719-1}
}

@inproceedings{Wuest_etal_2025,
  author = {Wüst, Antonia and Woydt, Tim and Helff, Lukas and Ibs, Inga
    and Stammer, Wolfgang and Dhami, Devendra S. and Rothkopf,
    Constantin A. and Kersting, Kristian},
  title = {Bongard in Wonderland: {{Visual}} Puzzles That Still Make
    {{AI}} Go Mad?},
  booktitle = {Proceedings of the 42nd international conference on
    machine learning (ICML)},
  year = {2025}
}

@inproceedings{Shafto_Goodman_2008,
  author = {Shafto, Patrick and Goodman, Noah},
  publisher = {Cognitive Science Society Austin, TX},
  title = {Teaching Games: {{Statistical}} Sampling Assumptions for
    Learning in Pedagogical Situations},
  booktitle = {Proceedings of the 30th annual conference of the
    cognitive science society},
  pages = {1632-1637},
  year = {2008}
}

@article{Shafto_etal_2014,
  author = {Shafto, Patrick and Goodman, Noah D. and Griffiths, Thomas
    L.},
  publisher = {Elsevier BV},
  title = {A Rational Account of Pedagogical Reasoning: {{Teaching}} by,
    and Learning from, Examples},
  journal = {Cognitive Psychology},
  volume = {71},
  pages = {55-89},
  year = {2014},
  month = {jun},
  url = {http://dx.doi.org/10.1016/j.cogpsych.2013.12.004},
  doi = {10.1016/j.cogpsych.2013.12.004}
}

@article{Depeweg_etal_2024,
  author = {Depeweg, Stefan and Rothkopf, Constantin A. and Jäkel,
    Frank},
  title = {Solving {{Bongard}} {{Problems}} {{With}} a {{Visual}}
    {{Language}} and {{Pragmatic}} {{Constraints}}},
  journal = {Cognitive Science},
  volume = {48},
  number = {5},
  pages = {e13432},
  year = {2024},
  month = {may},
  url = {https://onlinelibrary.wiley.com/doi/10.1111/cogs.13432},
  doi = {https://10.1111/cogs.13432}
}

@article{Flener_Schmid_2008,
  author = {Flener, Pierre and Schmid, Ute},
  title = {An Introduction to Inductive Programming},
  journal = {Artificial Intelligence Review},
  volume = {29},
  number = {1},
  pages = {45-62},
  year = {2008},
  month = {mar},
  url = {http://link.springer.com/10.1007/s10462-009-9108-7},
  doi = {10.1007/s10462-009-9108-7}
}

@book{Bongard_1970,
  author = {Bongard, Michail M.},
  editor = {Hawkins, Joseph K.},
  publisher = {Spartan Books},
  title = {Pattern Recognition},
  year = {1970},
  address = {New York}
}
\end{document}